\documentclass{svproc}
\usepackage{url}

\usepackage{graphicx}
\usepackage{tabularx}
\usepackage{adjustbox}
\usepackage{float}
\usepackage{amsmath}
\begin{document}
\mainmatter % % start of a contribution
\title{Multimodal RAG for Unstructured Data: Leveraging Modality-Aware Knowledge Graphs with Hybrid Retrieval}
\titlerunning{Multimodal RAG for Unstructured Data}  % abbreviated title (for running head)
\author{Rashmi R\inst{1} \and Vidyadhar Upadhya\inst{2}}
\authorrunning{Rashmi R et al.} % abbreviated author list (for running head)
\tocauthor{Vidyadhar Upadhya}
\institute{National Institute of Technology Karnataka, Surathkal, India,\\
\email{rashmir.243cd003@nitk.edu.in},
\and
National Institute of Technology Karnataka, Surathkal,\\India}

\maketitle              % typeset the title of the contribution

\begin{abstract}

Current Retrieval-Augmented Generation (RAG) systems primarily operate on unimodal textual data, limiting their effectiveness on unstructured multimodal documents. Such documents often combine text, images, tables, equations, and graphs, each contributing unique information. In this work, we present a Modality-Aware Hybrid retrieval Architecture (MAHA), designed specifically for multimodal question answering with reasoning through a modality-aware knowledge graph. MAHA integrates dense vector retrieval with structured graph traversal, where the knowledge graph encodes cross-modal semantics and relationships. This design enables both semantically rich and context-aware retrieval across diverse modalities. Evaluations on multiple benchmark datasets demonstrate that MAHA substantially outperforms baseline methods, achieving a ROUGE-L score of 0.486, providing complete modality coverage. These results highlight MAHA’s ability to combine embeddings with explicit document structure, enabling effective multimodal retrieval. Our work establishes a scalable and interpretable retrieval framework that advances RAG systems by enabling modality-aware reasoning over unstructured multimodal data.

\end{abstract}
\section{Introduction}
Efficiently retrieving and synthesizing information is paramount in an increasingly data-rich world, impacting decision-making in all fields. While some of this data exists in structured forms, a substantial and growing portion resides in unstructured formats \cite{UDA} including PDFs, scanned documents, raw text documents containing intricate tables, illustrative graphs, various image types, mathematical equations and extensive textual content.

The advent of Large Language Models (LLMs) has demonstrated unprecedented capabilities in tasks traditionally requiring human cognitive effort, including question answering, document summarization, and fact retrieval. At the forefront of this paradigm shift is Retrieval-Augmented Generation (RAG), a powerful technique that augments the generative capacity of LLMs by enabling them to fetch and incorporate relevant information from external data sources dynamically. This dynamic mechanism is crucial for mitigating \textquotedblleft hallucinations\textquotedblright (the generation of factually incorrect or fabricated information), thereby producing accurate and trustworthy responses.

In RAGs, the retriever plays a central role in identifying relevant information from large collections of documents. Different types of retrievers bring distinct strengths. Sparse retrievers, such as Best Matching-25 (BM25) \cite{10.1007/978-1-4471-2099-5_24}, rely on lexical overlap to locate documents containing exact query terms. They are efficient and effective for keyword-driven searches, often serving as an initial filtering step. However, because they lack semantic awareness, they perform poorly when queries involve synonyms or alternative phrases. Dense retrievers address this limitation by encoding queries and documents into high-dimensional embeddings using models like Sentence-BERT (SBERT)  \cite{sbert}. Retrieval then consists of finding semantically close vectors, typically with similarity search tools such as Facebook AI Similarity Search (FAISS) \cite{douze2025faisslibrary}. These methods excel at capturing meaning in natural language queries but can be less reliable when precise or rare terms are required. Hybrid retrievers combine sparse and dense techniques (e.g., BM25 with FAISS), integrating precision with semantic understanding to achieve more balanced and robust retrieval. For more complex queries that demand reasoning across multiple pieces of evidence, multi-hop or iterative retrievers are employed. These often leverage knowledge graphs (KGs) \cite{KG_IRAG}, where entities and their relationships are represented explicitly, enabling the system to traverse connected facts and synthesize comprehensive answers.

Current RAGs predominantly operate on unimodal textual corpora
and are limited when dealing with the complexities of unstructured documents that are inherently multimodal, where text is closely interlinked with images, tables, equations, and graphs, each adding a distinct semantic value. The challenge lies not only in extracting information from these varied modalities, but also in understanding and leveraging the intricate relationships between them. For example, a textual description might refer to data presented in a table, while a scientific finding might be visually represented by a graph or mathematically formalized by an equation. Traditional RAG systems often fail to capture these cross-modal dependencies, leading to incomplete retrieval and suboptimal generation.

Our work addresses these limitations with a novel multimodal and hybrid RAG framework for unstructured data comprising text, tables, images, graphs, and equations. The hybrid nature of our framework arises from (i) the fusion of dense vector and graph-based retrieval methods and (ii) the use of a modality-aware knowledge graph that integrates and links information across modalities. The nodes in the knowledge graph represent entities across different modalities (text, images, graphs, tables, equations), and edges capture their semantic and structural relationships, for example, how a table supports a paragraph or how an equation relates to a table, enabling deeper contextual understanding. The performance of the proposed hybrid RAG framework is systematically compared with a comprehensive set of established baseline retrievers to demonstrate its retrieval and generation quality. The evaluation uses a combination of quantitative and qualitative metrics to provide a holistic assessment of the performance of the framework.

\section{Prior Work, Limitations and Our Contributions}
Despite the progress in RAG, extending its capabilities to unstructured and multimodal data remains a significant challenge.
Several hybrid RAG frameworks integrate knowledge graphs with vector retrieval for enhanced information extraction \cite{HybridRAG,KG-guided-RAG,DO-RAG}.
 While these approaches improve accuracy and contextual relevance, their knowledge graphs are predominantly textual, lacking explicit support for multimodal inputs like graphs, tables, equations or images. This text-only focus limits their ability to capture and reason over cross-modal relationships inherent in many real-world documents.
There are attempts to retrieve information from semi-structured data and integrate diverse textual sources using hybrid RAG \cite{HybGRAG} and domain-specific KGs
\cite{KG-RAG-Schema}. However, these methods often treat non-textual elements as text fields or ignore them entirely. They lack robust mechanisms for multimodal reasoning, where the semantics of images, tables, or equations are fully integrated into the retrieval and generation process.

There exist RAGs for image retrieval; however, they fail if images lack relevant descriptions or if images are positioned far from their textual context \cite{Beyond_Text}. This indicates a broader challenge to achieve robust, context-sensitive image-text alignment beyond simple captions.

Large-scale multimodal RAG systems like Kosmos-1 \cite{kosmos}, MM-ReAct \cite{mmreact}, and multimodal chain-of-thought reasoning LLMs \cite{MM-CoT} demonstrate impressive unified vision-language understanding. However, they often operate in closed settings, limiting their adaptability to specific domains or user requirements. Furthermore, they frequently lack flexible graph-based rationales and do not provide fine-grained control over modality-aware retrieval or chunk-level semantics. This is crucial for explainability and precise information extraction in complex documents.

Existing approaches broadly fall short in five key areas.
\begin{enumerate}
 \item Insufficient cross-modal alignment, lacking robust mechanisms to semantically link content across diverse modalities.
 \item Limited structured reasoning, often failing to model the intricate interdependencies among multimodal components, which hinders their capacity for multi-step inference.
 \item Reliance on static retrieval processes yielding outdated or contextually irrelevant results, particularly in dynamic or evolving information spaces.
 \item Lack of tailored strategies to jointly handle text, images, tables, graphs, and equations in multimodal retrieval.
 \item Shallow or modality-agnostic KG integrations limit cross-modal reasoning and reduce interpretability.
\end{enumerate}

Several recent works have sought to enhance RAG by incorporating knowledge graphs or hybrid retrieval mechanisms. HybridRAG \cite{HybridRAG} combines dense vector retrieval with domain-specific financial KGs, but its entities are purely textual, preventing multimodal reasoning. HybGRAG \cite{HybGRAG} integrates vector and KG retrieval on semi-structured QA datasets, yet it does not cover unstructured data. Similarly, KG-Guided \cite{KG-guided-RAG} RAG expands retrieval diversity through KG-driven chunk re-ranking, but it lacks visual or tabular encoding modules. DO-RAG \cite{DO-RAG} dynamically constructs KGs from unstructured text and integrates vector retrieval for domain-specific electrical engineering documents and does not consider diagrams, schematics, and structured tables, restricting multimodal applicability. WeKnow-RAG \cite{xie2024weknowragadaptiveapproachretrievalaugmented} unifies sparse and dense retrieval with web search and KGs, supporting diverse textual sources, but lacks mechanisms for multimodal reasoning.

Our work addresses these limitations by developing a novel Modality-Aware Hybrid retrieval Architecture (MAHA) that supports structured and explainable reasoning over unstructured data. We propose a system integrating dense vector-based retrieval with structured graph traversal over a modality-aware knowledge graph. This integration combines the efficiency of vector-based similarity search with the precision and explainability of graph traversal. We demonstrate that the proposed system is capable of reasoning across modalities by leveraging both the semantic richness of embedding and the explicit structure of the knowledge graph on benchmark datasets.

\section{Modality-Aware Hybrid retrieval Architecture (MAHA)}
\subsection{Architecture}
\begin{figure}[ht]
 \includegraphics[width=\textwidth]{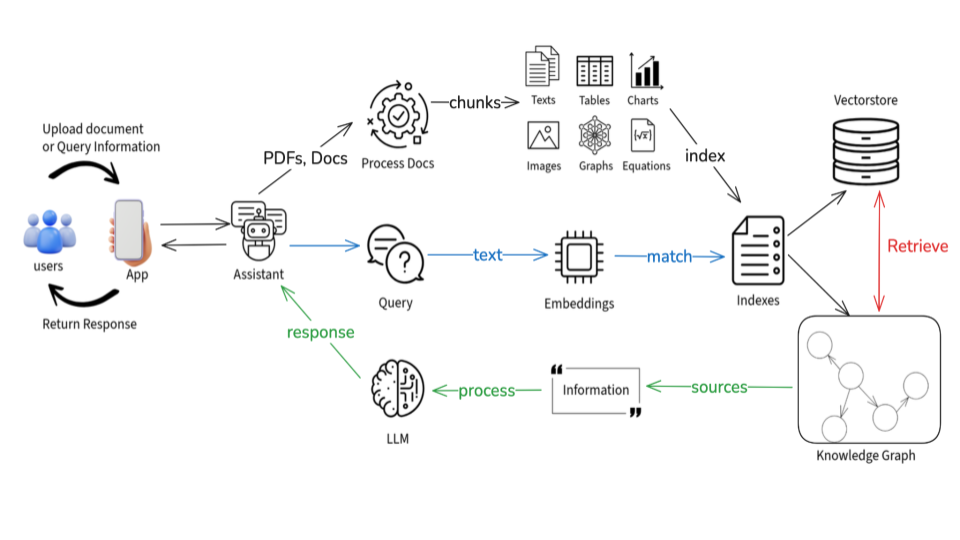}
 \caption{The proposed Modality-Aware Hybrid retrieval Architecture (MAHA).}\label{fig:arch}
\end{figure}
The proposed architecture is shown in Fig. \ref{fig:arch}.
The documents and queries submitted by the user are provided to the \textit{Assistant} module, which orchestrates the pipeline.
\subsubsection{Ingestion and Embedding:}
The documents are directed to the \textit{Process} module that performs multimodal document parsing, extracting and segmenting various content types, including text, tables, charts, images, graphs, and equations into semantically meaningful chunks. The text chunks are converted into embeddings using language models (e.g. OpenAI text-embedding-3-small), tables are converted to Hyper Text Markup Language (HTML) format, equations are encoded as structured equations (\LaTeX), and visual elements such as images and graphs are encoded using Contrastive Language-Image Pre-training model (CLIP: openai-clip-vit-base-patch32) \cite{clip} and converted to base64 format. Non-textual data is also summarized and embedded.
\subsubsection{Vectorstore Indexing and Knowledge Graph Construction:}
\begin{figure}[ht]
    \centering
    \includegraphics[width=.6\textwidth]{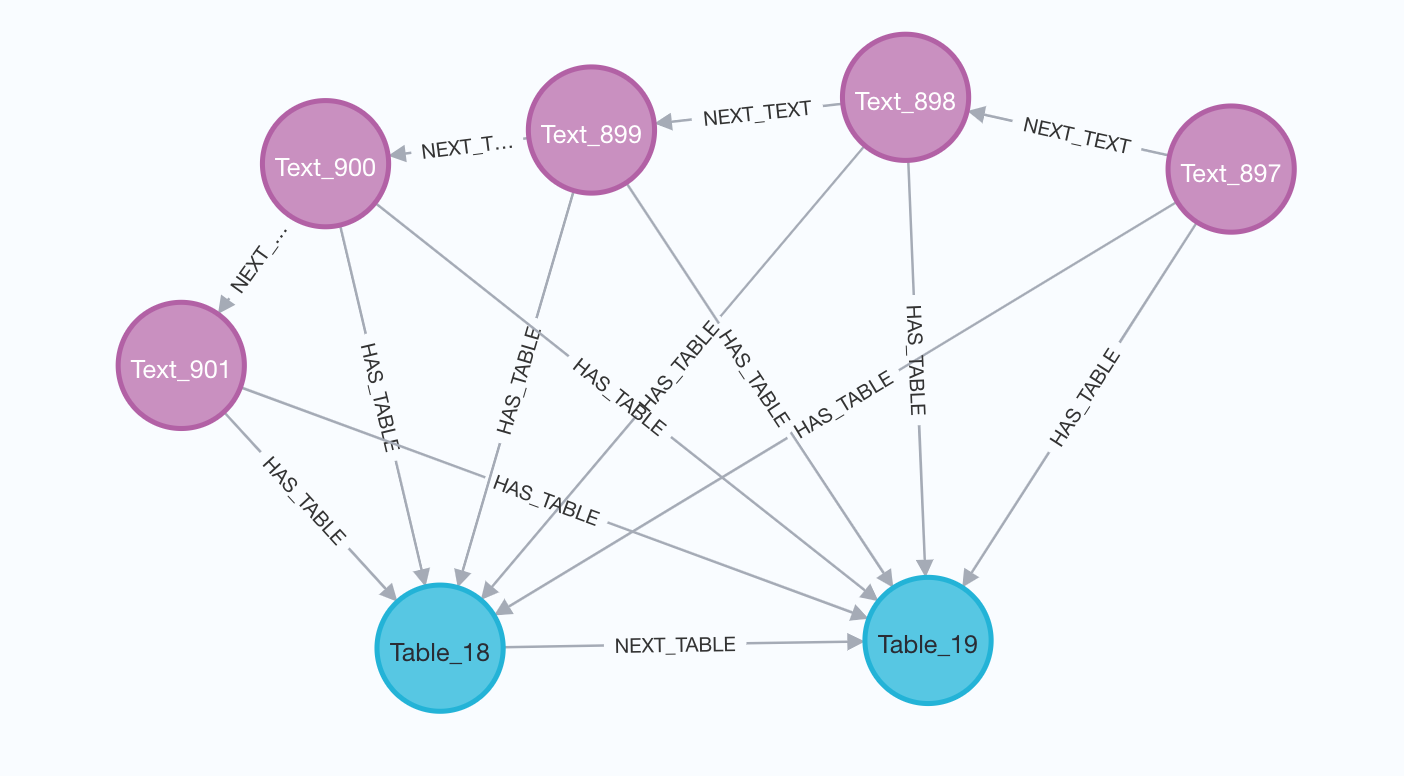}
    \includegraphics[width=.6\textwidth]{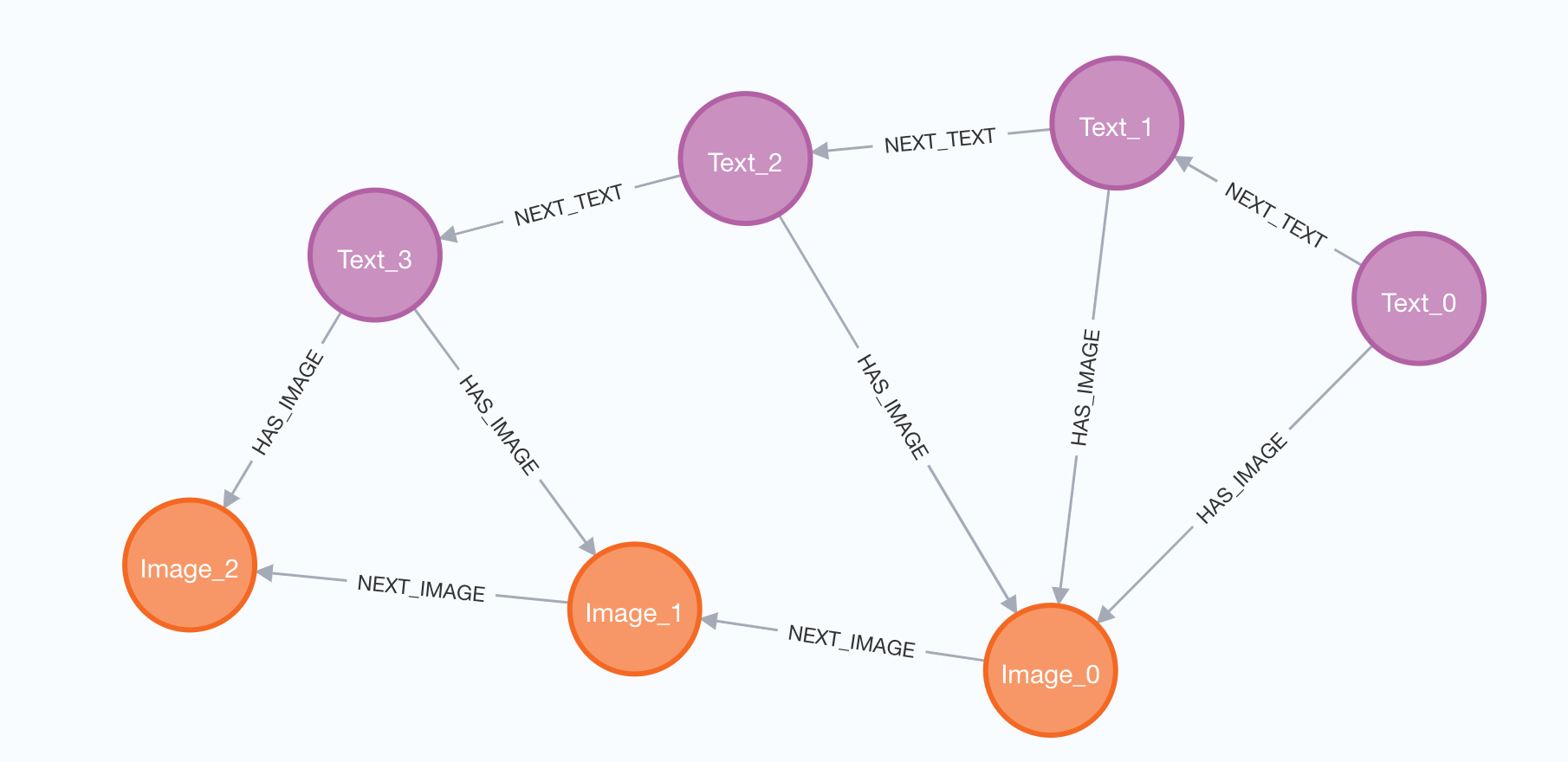}
    \caption{Text and Table, Text and Image Relationships captured through KGs}\label{fig:KG}
\end{figure}

The representations are then indexed into a vectorstore, allowing fast similarity-based retrieval across modalities. Along with the vectorstore a knowledge graph is built to capture the relationships between embeddings. Existing schemas are largely text-centric; we extended them with modality-aware relationships to capture cross-modal semantics. The nodes in the graph represent entities such as text, equations, images, and tables. The edges in the graph capture semantic relationships such as \textquotedblleft NEXT - TEXT\textquotedblright, \textquotedblleft NEXT - TABLE\textquotedblright, \textquotedblleft NEXT - IMAGE\textquotedblright, \textquotedblleft NEXT - FORMULA\textquotedblright, \textquotedblleft HAS - IMAGE\textquotedblright, \textquotedblleft HAS -  TABLE\textquotedblright, and \textquotedblleft HAS - FORMULA\textquotedblright, as demonstrated in Fig. \ref{fig:KG}. This graph provides a structure to the data and supports reasoning over the retrieved data. Graph construction is schema-driven and includes named entity linking, coreference resolution, and relationship inference.
\subsubsection{Hybrid Retrieval (Vector + Graph):}
When a user submits a query, the \textit{Query} module encodes it into embeddings via a text-to-vector transformation. These embeddings are matched against the indexed chunks in the vectorstore to retrieve semantically similar content. Simultaneously, the knowledge graph is queried to retrieve supporting information based on entity relations and graph traversal. A key challenge was balancing semantic similarity with structural traversal; we designed a fusion strategy to integrate both without sacrificing relevance or coverage. The indexes serve as the common link between semantic and structured retrieval. Combining these approaches ensures both modality coverage and contextual depth, especially in cases where answers are spread across related sections or modalities.
\subsubsection{Context-Aware Generation with LLMs:}
The retrieved content is passed to an \textit{LLM} to synthesize the information into a coherent response. The LLM uses prompts that include query context, retrieved evidence, and graph metadata to generate accurate, explainable, and contextually grounded answers.

\section{Experimental Setup}
\subsection{Datasets}
The effectiveness of our proposed framework (MAHA) is validated through experiments on the following benchmark datasets.
\begin{itemize}
\item \textbf{Unstructured Document Analysis (UDA) Benchmark Suite:
\cite{UDA}}
contains data from the following domains.
\begin{itemize}
\item \textbf{Financial} domain data comprises of financial reports with intricate layouts, including text, image and tabular data, challenging the system with numerical reasoning tasks.
\item The data from \textbf{Academia} is sourced from academic papers and it tests the system's ability to reason over complex technical content, graphs, tables and equations.
\item The world knowledge data from \textbf{Wikipedia} includes a mix of text, image and tabular data from Wikipedia pages, evaluating performance on a wide variety of topics.
\end{itemize}
\item \textbf{MRAMG-Bench:\cite{mram}} This dataset is extracted from web, academia and lifestyle domains that include text, images, graphs, tables and equations. This benchmark is specifically designed in \cite{mram} to test multimodal reasoning capabilities, requiring models to integrate information to generate a single, coherent answer.
\item \textbf{REAL-MM-RAG-Bench: \cite{Real-MM-RG}} This is a high-quality
benchmark dataset curated in \cite{Real-MM-RG}, containing text, tables, and images from the financial domain, to validate the multimodal question-answering tasks.
\end{itemize}
% These datasets collectively provide a rigorous testing ground for our framework, allowing us to assess its performance on a wide range of real-world, multimodal challenges.
\subsection{Baseline  Retrieval Systems}
We consider the following baseline models for comparison.
\begin{itemize}
\item \textbf{BM25 (Sparse Retriever)}: A classical lexical retrieval model based on term frequency and inverse document frequency, limited to exact keyword matches without semantic or multimodal understanding  
\item \textbf{FAISS + SBERT (Dense Retriever)}: A vector-based similarity search over dense embeddings and its shortcomings in capturing logical connections between modalities.  
\item \textbf{CLIP (Image-only Retriever)}: This model serves as a key baseline for image-centric questions, highlighting the image understanding component of our framework.
\item \textbf{Hybrid (BM25 + FAISS)}: A common ensemble approach serving as a strong hybrid baseline to assess the added value of the knowledge graph in our framework.
\item \textbf{Graph Traversal (KG Retriever)}: A structural retrieval method leveraging knowledge graphs to capture contextual linkages, but limited in semantic depth and multimodal coverage.
\end{itemize}
\subsection{Evaluation Metrics}
The performance of the retrieval systems is evaluated using the following metrics.
\begin{itemize}
 \item \textbf{Retrieval Metrics:}
\begin{itemize}
    \item \textbf{Recall@K}: It measures the proportion of queries for which the correct document chunk is among the top-$k$ retrieved results.
%     This is critical for evaluating the system's ability to find the necessary information.
    \item \textbf{Mean Reciprocal Rank (MRR)}: It measures the rank of the first correct answer. A higher MRR indicates that the system is not only finding the correct information but also ranking it highly.
\end{itemize}
\item \textbf{Generation Metrics:}
\begin{itemize}
    \item \textbf{ROUGE-$L$}: It measures the overlap of the longest common subsequence between the generated answer and the ground truth answer. This metric will assess the factual accuracy and completeness of the generated response.
\end{itemize}
\item \textbf{Multimodal Metric:}
\begin{itemize}
    \label{reference: modality-coverage}
    \item \textbf{Modality Coverage:} We define a new metric to capture the ability of a retrieval system to incorporate evidence across modalities. For each query $q$, let $M_{gt}(q)$ be the set of modalities required in the ground truth answer, and $M_{ret}(q)$ the set of modalities retrieved by the system. The coverage for the query $q$ is:
    \[
    \text{Coverage}(q) = \frac{|M_{gt}(q) \cap M_{ret}(q)|}{|M_{gt}(q)|}.
    \]
    The overall Modality Coverage is the average across all queries:
    \[
    \text{Modality Coverage} = \frac{1}{N} \sum_{i=1}^{N} \text{Coverage}(q_i).
    \]
    A score of $1.0$ indicates that the system consistently retrieves all required modalities, while lower scores reflect partial retrieval.
\end{itemize}
\end{itemize}
\subsection{Challenges}
Building the proposed RAG framework (MAHA) introduced several practical challenges:
\begin{itemize}
    \item \textbf{Parsing heterogeneous content:} Unlike textual inputs, multimodal documents lack a consistent structure. Segmenting images, tables, and equations into meaningful chunks and aligning them with related text required tailored strategies.
    \item \textbf{Extending Knowledge Graph schemas:} Most KG designs are text-centric. We had to introduce modality-aware relationships (e.g., HAS-IMAGE, HAS-TABLE, NEXT-FORMULA) to capture cross-modal semantics.
    \item \textbf{Coordinating hybrid retrieval:} Balancing when to prioritize semantic vector retrieval versus graph traversal required a fusion strategy that preserved both relevance and multimodal coverage.
\end{itemize}

\section{Experimental Results}
This section presents the results of our experiments, comparing the performance of MAHA against the baseline retrieval systems and KG based multimodal RAG frameworks.
\subsection{MAHA vs Baseline Retrieval Systems}

\begin{figure}[H]
    \centering
    \includegraphics[width=\textwidth]{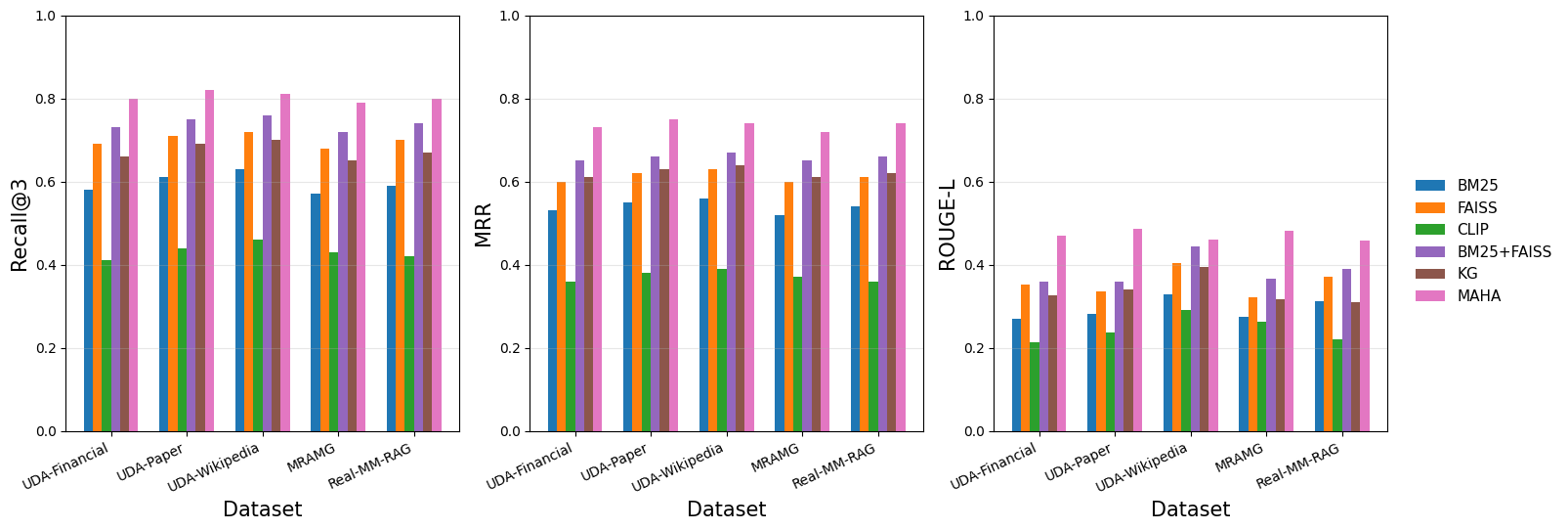}
    \caption{Performance of MAHA compared with baseline retrievers.}
    \label{fig:baseline-barchart}
\end{figure}
% Figure \ref{fig:all-baseline-retrievers} presents a comparative performance analysis of various retrieval systems across five different multimodal datasets: UDA-Financial, UDA-Paper, UDA-Wikipedia, MRAMG, and Real-MM-RAG. The results are evaluated using three key metrics: Recall@3, MRR (Mean Reciprocal Rank), and ROUGE-L.
Figure \ref{fig:baseline-barchart} clearly shows that MAHA consistently outperforms all other baseline models, demonstrating its superior ability to retrieve and rank relevant information. It achieves the highest scores across all three metrics and datasets. In contrast, single-modality retrievers like BM25, CLIP and FAISS show lower performance on complex multimodal documents. While the hybrid BM25+FAISS model does improve performance over the individual BM25 and FAISS models, it still falls short compared to MAHA.
\subsection{MAHA vs Multimodal RAG Frameworks}
To validate the efficiency of MAHA, we compare it with the existing Multimodal RAG frameworks that use KG integration.
\begin{figure}[H]
    \centering
    \includegraphics[width=\textwidth]{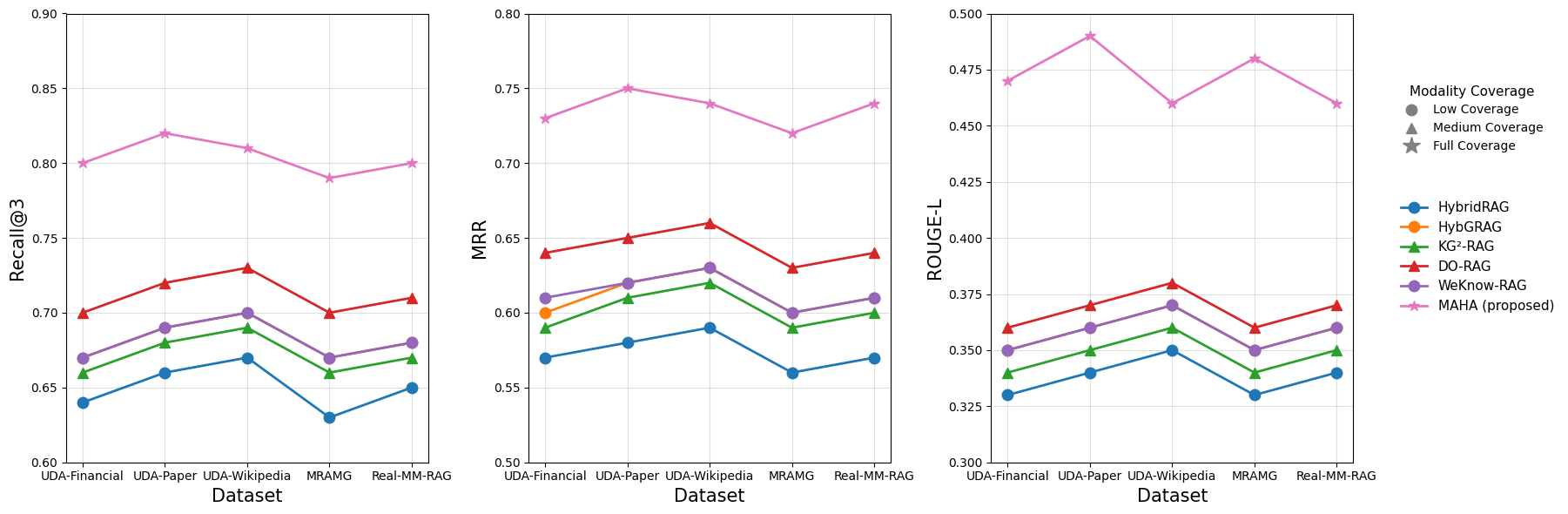}
    \caption{Performance and Modality Coverage of MAHA compared with KG-based multimodal RAG frameworks.}
\label{fig: paper_comparision}
\end{figure}
In Fig. \ref{fig: paper_comparision}, the performance is reported across Recall@3, MRR, and ROUGE-L, with marker shapes indicating modality coverage. We observe that MAHA consistently outperforms all other existing systems across all three metrics. It achieves the highest ROUGE-L score (0.486), the highest Recall@3 (0.81), and the highest MRR (0.74), while also being the only method to achieve full modality coverage (1.00). In contrast, prior works like HybridRAG, HybGRAG, Knowledge Graph-Guided RAG, DO-RAG, and WeKnow-RAG show lower performance and limited modality coverage (0.00-0.39).
\subsection{Ablation Study}
To quantify the contribution of each retrieval component and validate our design choices, we performed controlled ablation experiments. Specifically, we compared (i) a purely vector-based retriever using dense embeddings (Vector-Only), (ii) a purely graph-based retriever relying solely on structural traversal (Graph-Only), and (iii) MAHA, integrating dense retrieval with a modality-aware knowledge graph.
\begin{figure}[H]
    \centering
    \includegraphics[width=.8\textwidth]{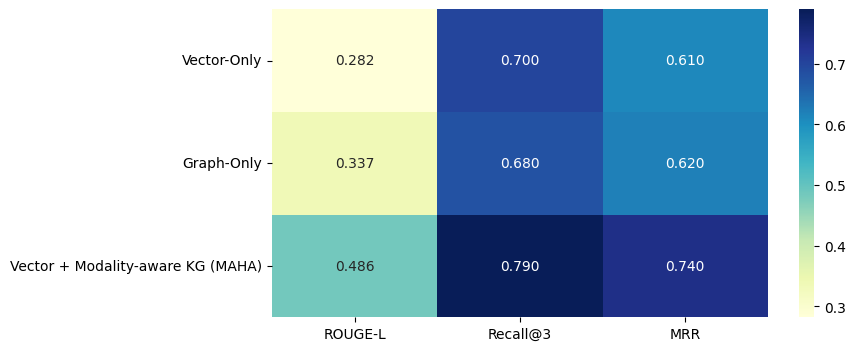}
    \caption{Performance of MAHA compared with vector-only and graph-only retrievers.}
    \label{fig:ablation_studies}
\end{figure}
From the Fig. ~\ref{fig:ablation_studies}, we observe that the Vector-Only baseline achieves moderate accuracy (ROUGE-L: $0.282$, Recall@3: $0.70$, MRR: $0.61$), confirming that dense representations capture local semantics but fail to account for structural and cross-modal cues. The graph-only retriever improves answer quality (ROUGE-L $0.337$) and maintains comparable retrieval effectiveness (Recall@3 $0.68$, MRR $0.62$), demonstrating that structural relations between content units provide complementary information. This suggests that structural navigation aids ranking but struggles to surface rich evidence on its own.

The proposed MAHA delivers substantial improvements: ROUGE-L $0.486$, Recall@3 $0.79$, and MRR $0.74$. This represents a relative gain of approximately \textbf{72\%} in ROUGE-L over the vector-only baseline, and \textbf{44\%} over the graph-only variant. The Recall@3 increase from $0.70$ (vector-only) and $0.68$ (graph-only) to $0.79$ demonstrates broader retrieval coverage, while the MRR gain from $0.61$/$0.62$ to $0.74$ highlights an improvement of around \textbf{21\%} and \textbf{19\%}, respectively, in ranking quality, while achieving full modality coverage.

These results highlight two key insights: (1) structural reasoning in the KG complements semantic similarity from dense vectors, and (2) explicit modality-aware links in the KG (e.g., HAS-IMAGE, HAS-TABLE) enable retrieval of multimodal evidence that would otherwise be missed.
\section{Conclusion and Future Work}
In this work, we have successfully developed and validated a novel multimodal RAG architecture: Modality-Aware Hybrid retrieval Architecture (MAHA). By strategically integrating a modality-aware knowledge graph with a vector-based indexing technique, we have addressed and overcome the significant challenges inherent in reasoning across complex cross-modal data within documents. Our hybrid retrieval system moves beyond the limitations of traditional RAG systems, which often fail to capture the logical and relational nuances of complex information.

The results of our comprehensive experimental evaluation and detailed ablation studies prove the efficacy of our approach. MAHA did not merely improve upon existing methods; it achieved a substantial and conclusive gain in key metrics, including ROUGE-L, Recall@3, and MRR. The ablation analysis confirmed our core hypothesis. This performance leap is not due to any single component but is a direct result of the powerful synergy between semantic (vector) and relational (graph) retrieval.

While the model demonstrates a robust foundation, we believe there is significant potential in exploring more advanced, automated methods for knowledge graph construction to handle highly unstructured data formats. Additionally, future research could focus on developing a more dynamic query router that can intelligently adapt to the complexity of a user's question in real-time.

In summary, we are confident that this research provides an intelligent solution for multimodal information retrieval, setting a new benchmark for RAG systems aiming to unlock the full potential of complex, multimodal data.

\bibliographystyle{plain}
\bibliography{references}
% \nocite{*}
% \begin{thebibliography}{6}
% %

% \bibitem {smit:wat}
% Smith, T.F., Waterman, M.S.: Identification of common molecular subsequences.
% J. Mol. Biol. 147, 195?197 (1981). \url{doi:10.1016/0022-2836(81)90087-5}

% \bibitem {may:ehr:stein}
% May, P., Ehrlich, H.-C., Steinke, T.: ZIB structure prediction pipeline:
% composing a complex biological workflow through web services.
% In: Nagel, W.E., Walter, W.V., Lehner, W. (eds.) Euro-Par 2006.
% LNCS, vol. 4128, pp. 1148?1158. Springer, Heidelberg (2006).
% \url{doi:10.1007/11823285_121}

% \bibitem {fost:kes}
% Foster, I., Kesselman, C.: The Grid: Blueprint for a New Computing Infrastructure.
% Morgan Kaufmann, San Francisco (1999)

% \bibitem {czaj:fitz}
% Czajkowski, K., Fitzgerald, S., Foster, I., Kesselman, C.: Grid information services
% for distributed resource sharing. In: 10th IEEE International Symposium
% on High Performance Distributed Computing, pp. 181?184. IEEE Press, New York (2001).
% \url{doi: 10.1109/HPDC.2001.945188}

% \bibitem {fo:kes:nic:tue}
% Foster, I., Kesselman, C., Nick, J., Tuecke, S.: The physiology of the grid: an open grid services architecture for distributed systems integration. Technical report, Global Grid
% Forum (2002)

% \bibitem {onlyurl}
% National Center for Biotechnology Information. \url{http://www.ncbi.nlm.nih.gov}

% \end{thebibliography}
\end{document}